# Semi-Analytic Method for SINS Attitude and Parameters Online Estimation

Lubin Chang

Department of Navigation Engineering, Naval University of Engineering, Wuhan, China

changlubin@163.com

*Abstract*—In this note, the attitude and inertial sensors drift biases estimation for Strapdown inertial navigation system is investigated. A semi-analytic method is proposed, which contains two interlaced solution procedures. Specifically, the attitude encoding the body frame changes and gyroscopes drift biases are estimated through attitude estimation while the attitude encoding the constant value at the very start and accelerometers drift biases are determined through online optimization.

## I. Introduction

The past ten years have seen the prosperity of initial alignment methods development for the strapdown inertial navigation system (SINS). This is mainly promoted by the optimization based alignment (OBA) methods which are proposed and popularized by Wu, Yan, et al. [1-9]. As the greatest contribution, OBA formulates the SINS attitude alignment into an attitude determination problem using vector observations, which "*reveals an interesting link between these two individual problems that has been studied in parallel for several decades* [1]". In OBA, the attitude is decomposed into separate Earth motion, inertial rate, and constant attitude at the very start. The attitude determination just refers to the determination of the constant attitude at the very start. Within the OBA framework, there are two main research aspects, that is, construction of the vector observations based on specific forces equation and extension of attitude determination model through formulating parameters other than attitude into the cost function for optimization. Regarding to the first aspect, Silson proposes an interleaved vector observations construction method, which can make full use of information inherent in the specific forces equation [9]. This procedure has also been used in the attitude estimation based alignment that will be discussed in the following right now. Regarding to the second aspect, Chang et al. formulate the attitude alignment into an attitude estimation method, which can estimate the gyroscopes drift biases coupled with the attitude [10, 11]. However, this method is powerless in estimating accelerometers drift biases. In contrast, Wu et al. propose an online constrained optimization method which can simultaneously estimate the attitude and other related parameters [4]. From the numerical test results, it is shown that this method is only validate in estimating accelerometers drift biases and is compromised in estimating gyroscopes drift biases. It is therefore desired to combine the advantages of Wu and Chang's methods, which is just the consideration of the presented work. In this note, a semi-analytic alignment (SAA) method is proposed, which is intended to realizing the aforementioned combination. In the proposed SAA, the attitude encoding body frame changes and gyroscopes drift biases are estimated using attitude estimation methods as in [10] and constant attitude at the very start and accelerometers drift biases are determined through optimization based method as in [4]. The attitude estimation and optimization are interlaced with each other in the SAA framework.

The contents are organized as follows. Section II mathematically formulates the SAA framework. Section III presents the solution procedures for the two interlaced sub-problems in SAA. Conclusions are drawn in Section IV. Finally, some of my comments on existing initial alignment methods are presented in Section V.

## II. Problem Formulation

All the quantities involved in this note uses the same set of symbols as in [4] for brevity and, if not stated, their definitions and meanings are omitted for brevity.

The attitude matrix $\mathbf{C}_b^n(t)$ can be decomposed as

$$\mathbf{C}_b^n(t) = \mathbf{C}_{n(0)}^{n(t)} \mathbf{C}_b^n(0) \mathbf{C}_{b(t)}^{b(0)} \tag{1}$$

where $\mathbf{C}_{n(0)}^{n(t)}$ can be calculated according to

$$\dot{\mathbf{C}}_{n(t)}^{n(0)} = \mathbf{C}_{n(t)}^{n(0)} \left( \boldsymbol{\omega}_{in}^n \times \right) \tag{2}$$

$\mathbf{C}_b^n(0)$ can be determined according to the procedures in [4] as

$$\mathbf{C}_b^n(0)(\boldsymbol{\alpha} + \boldsymbol{\chi}\mathbf{b}_a) = \boldsymbol{\beta} \tag{3}$$

where

$$\boldsymbol{\alpha} = \int_{t_m}^{t} \mathbf{C}_{b(\tau)}^{b(0)} \mathbf{f}^b d\tau = \sum_{k=m}^{M-1} \mathbf{C}_{b(t_k)}^{b(0)} \left( \Delta t \mathbf{I}_3 + \frac{\Delta t^2}{2} \boldsymbol{\omega}_{ib}^b \times \right) \mathbf{f}^b \tag{4}$$

$$\boldsymbol{\chi} = \int_{t_m}^{t} \mathbf{C}_{b(\tau)}^{b(0)} d\tau = \sum_{k=m}^{M-1} \mathbf{C}_{b(t_k)}^{b(0)} \left( \Delta t \mathbf{I}_{3\times 3} + \frac{\Delta t^2}{2} \left( \boldsymbol{\omega}_{ib}^b \times \right) \right) \tag{5}$$

$$\boldsymbol{\beta} = \mathbf{C}_{n(t)}^{n(0)} \mathbf{v}^n(t) - \mathbf{C}_{n(t_m)}^{n(0)} \mathbf{v}^n(t_m) + \int_{t_m}^{t} \mathbf{C}_{n(\tau)}^{n(0)} \boldsymbol{\omega}_{ie}^n \times \mathbf{v}^n d\tau - \int_{t_m}^{t} \mathbf{C}_{n(\tau)}^{n(0)} \mathbf{g}^n d\tau \tag{6}$$

The first integral on the right hand of (6) is given by

$$\int_{t_m}^{t} \mathbf{C}_{n(\tau)}^{n(0)} \boldsymbol{\omega}_{ie}^n \times \mathbf{v}^n d\tau = \sum_{k=m}^{M-1} \mathbf{C}_{n(t_k)}^{n(0)} \int_{t_k}^{t_{k+1}} \mathbf{C}_{n(\tau)}^{n(t_k)} \boldsymbol{\omega}_{ie}^n \times \mathbf{v}^n d\tau$$

$$\approx \sum_{k=m}^{M-1} \mathbf{C}_{n(t_k)}^{n(0)} \left[ \left( \frac{\Delta t}{2} \mathbf{I}_{3\times 3} + \frac{\Delta t^2}{6} \left( \boldsymbol{\omega}_{in}^n \times \right) \right) \boldsymbol{\omega}_{ie}^n \times \mathbf{v}^n(t_k) + \left( \frac{\Delta t}{2} \mathbf{I}_{3\times 3} + \frac{\Delta t^2}{3} \left( \boldsymbol{\omega}_{in}^n \times \right) \right) \boldsymbol{\omega}_{ie}^n \times \mathbf{v}^n(t_{k+1}) \right] \tag{7}$$

The second integral on the right hand of (6) is given by

$$\int_{t_m}^{t} \mathbf{C}_{n(\tau)}^{n(0)} \mathbf{g}^n d\tau = \sum_{k=m}^{M-1} \mathbf{C}_{n(t_k)}^{n(0)} \int_{t_k}^{t_{k+1}} \mathbf{C}_{n(\tau)}^{n(t_k)} \mathbf{g}^n d\tau \approx \sum_{k=m}^{M-1} \mathbf{C}_{n(t_k)}^{n(0)} \left( \Delta t \mathbf{I}_{3\times 3} + \frac{\Delta t^2}{2} \left( \boldsymbol{\omega}_{in}^n \times \right) \right) \mathbf{g}^n \tag{8}$$

In (3) the accelerometers drift biases $\mathbf{b}_a$ has also been jointly estimated associated with the attitude.

$\mathbf{C}_{b(t)}^{b(0)}$ and the gyroscopes drift biases $\mathbf{b}_g$ can be estimated according to the procedures in [10] as

$$\dot{\mathbf{C}}_{b(t)}^{b(0)} = \mathbf{C}_{b(t)}^{b(0)} \left( \tilde{\boldsymbol{\omega}}_{ib}^b - \mathbf{b}_g - \boldsymbol{\eta}_{gv} \right) \times \tag{9}$$

$$\mathbf{C}_{b(t)}^{b(0)} \boldsymbol{\alpha}_m = \boldsymbol{\beta}_m \tag{10}$$

where (9) is the process model and (10) is the measurement model. The vector observations in (10) are given by

$$\boldsymbol{\alpha}_m = \int_{t_m}^{t} \mathbf{C}_{b(0)}^{b(t)} \mathbf{C}_{b(\tau)}^{b(0)} \mathbf{f}^b(\tau) d\tau = \mathbf{C}_{b(0)}^{b(t)} \int_{t_m}^{t} \mathbf{C}_{b(\tau)}^{b(0)} \mathbf{f}^b(\tau) d\tau \tag{11}$$

$$\boldsymbol{\beta}_m = \mathbf{C}_b^n(0)^T \boldsymbol{\beta} \tag{12}$$

The integration in (11) can be calculated according to (4).

The estimation of $\mathbf{C}_{b(t)}^{b(0)}$ necessitates the constant attitude $\mathbf{C}_b^n(0)$ to construct the vector observations as shown in (12). The determination of $\mathbf{C}_b^n(0)$ necessitates the attitude $\mathbf{C}_{b(t)}^{b(0)}$ to construct the vector observations as shown in (4). This is just an interlaced procedure as shown in Fig. 1.

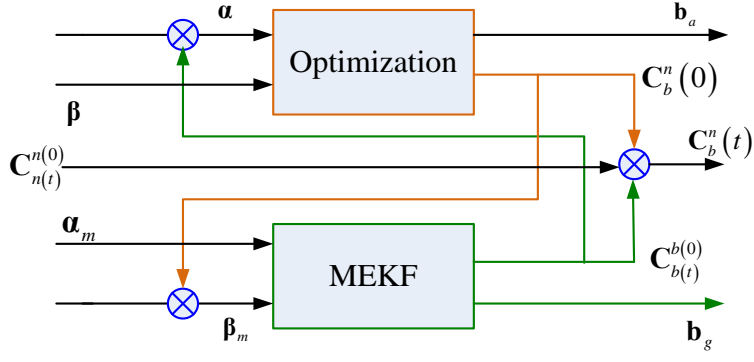

Fig. 1. Interlaced framework of SAA

## III. Problem Solutions

Through the formulation in last section, the attitude alignment and parameters estimation have been decomposed into two interlaced sub-problems. One refers to constrained-optimization and the other refers to attitude estimation. The solution procedures can be derived directly through lessons learned from the references [4, 12].

The constrained-optimization problem (3) is actually a simplified form of (27) in [4]. In this respect, the time-recursive Newton-Lagrange algorithm developed in [4] can be readily used to solve (3). However, due to the simplicity of (3), an efficient solution learned from the *Point-Cloud Alignment Problem* for robotics can be applied directly. The following procedures are the representation of that in section 8.1 of [13], where only notation modifications are made to adapt to problem (3).

Problem (3) can be reorganized as

$$\underbrace{\begin{bmatrix} \mathbf{C}_b^n(0) & 0 \\ \mathbf{0}_{1\times 3} & 1 \end{bmatrix}}_{(\mathbf{q}^{-1})^+ \mathbf{q}^\oplus} \left( \underbrace{\begin{bmatrix} \boldsymbol{\alpha} \\ 1 \end{bmatrix}}_{\mathbf{p}_j} + \underbrace{\begin{bmatrix} \chi \mathbf{b}_a \\ 0 \end{bmatrix}}_{\mathbf{r}} \right) = \underbrace{\begin{bmatrix} \boldsymbol{\beta} \\ 1 \end{bmatrix}}_{\mathbf{y}_j} \tag{13}$$

where $\mathbf{q}$ is the attitude quaternion corresponding to $\mathbf{C}_b^n(0)$ with definition as $\mathbf{q} = \begin{bmatrix} \boldsymbol{\rho}^T & \eta \end{bmatrix}^T$. The quaternion left-hand compound operator "+" and the right-hand compound operator "$\oplus$" are defined as

$$\mathbf{q}^+ = \begin{bmatrix} \eta \mathbf{I}_{3\times 3} - (\boldsymbol{\rho}\times) & \boldsymbol{\rho} \\ -\boldsymbol{\rho}^T & \eta \end{bmatrix} \tag{14}$$

$$\mathbf{q}^\oplus = \begin{bmatrix} \eta \mathbf{I}_{3\times 3} + (\boldsymbol{\rho}\times) & \boldsymbol{\rho} \\ -\boldsymbol{\rho}^T & \eta \end{bmatrix} \tag{15}$$

Given the corresponding definitions in the subscripts of (13), an error quaternion can be formed as

$$\mathbf{e}_j = \mathbf{y}_j - \left(\mathbf{q}^{-1}\right)^{+} \left(\mathbf{p}_j + \mathbf{r}\right)^{+} \mathbf{q} \tag{16}$$

The total objective function to be minimized can be defined as

$$J(\mathbf{q},\mathbf{r},\lambda) = \frac{1}{2}\sum_{j=1}^{M} w_j \mathbf{e}_j^T \mathbf{e}_j - \frac{1}{2}\lambda\left(\mathbf{q}^T\mathbf{q}-1\right) \tag{17}$$

where $w_j$ is the weight corresponding to $\mathbf{e}_j$. Taking the derivative of the objective function with respect to $\mathbf{q}$, $\mathbf{r}$ and $\lambda$ gives

$$\frac{\partial J}{\partial \mathbf{q}^T} = \sum_{j=1}^{M} w_j \left(\mathbf{y}_j^{\oplus} - \left(\mathbf{p}_j+\mathbf{r}\right)^{+}\right)^T \left(\mathbf{y}_j^{\oplus} - \left(\mathbf{p}_j+\mathbf{r}\right)^{+}\right)\mathbf{q} - \lambda\mathbf{q} \tag{18}$$

$$\frac{\partial J}{\partial \mathbf{r}^T} = -\mathbf{q}^{-1\oplus}\sum_{j=1}^{M} w_j \left(\mathbf{y}_j^{\oplus} - \left(\mathbf{p}_j+\mathbf{r}\right)^{+}\right)\mathbf{q} \tag{19}$$

$$\frac{\partial J}{\partial \lambda} = -\frac{1}{2}\left(\mathbf{q}^T\mathbf{q}-1\right) \tag{20}$$

Setting (19) to zero, we find

$$\mathbf{r} = \mathbf{q}^{+}\mathbf{y}^{+}\mathbf{q}^{-1} - \mathbf{p} \tag{21}$$

where $\mathbf{y}$ and $\mathbf{p}$ are given in the following. Substituting (21) into (18) and setting the resulting equation to zero, we have

$$\mathbf{W}\mathbf{q} = \lambda\mathbf{q} \tag{22}$$

where

$$\mathbf{W} = \frac{1}{w}\sum_{j=1}^{M} w_j \left(\left(\mathbf{y}_j-\mathbf{y}\right)^{\oplus} - \left(\mathbf{p}_j-\mathbf{p}\right)^{+}\right)^T \left(\left(\mathbf{y}_j-\mathbf{y}\right)^{\oplus} - \left(\mathbf{p}_j-\mathbf{p}\right)^{+}\right) \tag{23}$$

$$\mathbf{y} = \frac{1}{w}\sum_{j=1}^{M} w_j \mathbf{y}_j, \quad \mathbf{p} = \frac{1}{w}\sum_{j=1}^{M} w_j \mathbf{p}_j, \quad w = \sum_{j=1}^{M} w_j \tag{24}$$

It is shown that (22) is just an eigenvalue problem, which can be solved using some existing well-known methods, such as Davenport q-method as in [1]. The accelerometer drift bias $\mathbf{b}_a$ can be readily identified from (21).

The attitude estimation problem (9) and (10) can be solved using the representative algorithms, such as multiplicative extended Kalman filter (MEKF) or UnScented QUaternion Estimator (USQUE) [13]. If the constant attitude $\mathbf{C}_b^n(0)$ is initialized using the eigenvalue problem (3) ignoring sensors biases, the MEKF will be adequate for the attitude estimation problem (9) and (10). This is because that the initial value of the attitude is accurately known, that is $\mathbf{C}_{b(0)}^{b(0)} = \mathbf{I}_{3\times3}$. The explicit procedures for estimation of $\mathbf{C}_{b(t)}^{b(0)}$ coupled with $\mathbf{b}_g$ by MEKF are summarized in **Algorithm 1**.

---

**Algorithm 1**: MEKF for estimation of $\mathbf{C}_{b(t)}^{b(0)}$ and $\mathbf{b}_g$

---

**Initialization**

$\hat{\mathbf{q}}_{b_0}^{b}(t_0) = [0\ 0\ 0\ 1]^T$, $\mathbf{b}_g(t_0) = \mathbf{0}_{3\times 1}$ and $\mathbf{P}(t_0) = \mathbf{P}_0$

**Kalman Gain**

$$\mathbf{K}_k = \mathbf{P}_{k|k-1}\mathbf{H}_k^T(\hat{\mathbf{x}}_{k|k-1})\left[\mathbf{H}_k(\hat{\mathbf{x}}_{k|k-1})\mathbf{P}_{k|k-1}\mathbf{H}_k^T(\hat{\mathbf{x}}_{k|k-1}) + \mathbf{R}_k\right]^{-1}$$

$$\mathbf{H}_k(\hat{\mathbf{x}}_{k|k-1}) = \left[\mathbf{C}(\hat{\mathbf{q}}_{b_0,k|k-1}^b)\boldsymbol{\beta}_{m,k}\quad \mathbf{0}_{3\times 3}\right]$$

**Update**

$$\mathbf{P}_k = \left[\mathbf{I}_{6\times 6} - \mathbf{K}_k\mathbf{H}_k(\hat{\mathbf{x}}_{k|k-1})\right]\mathbf{P}_{k|k-1}$$

$$\Delta\hat{\mathbf{x}}_k = \mathbf{K}_k\left[\boldsymbol{\alpha}_{m,k} - h_k(\hat{\mathbf{x}}_{k|k-1})\right]$$

$$h_k(\hat{\mathbf{x}}_{k|k-1}) = \mathbf{C}(\hat{\mathbf{q}}_{b_0,k|k-1}^b)\boldsymbol{\beta}_{m,k}$$

$$\Delta\hat{\mathbf{x}}_k = \left[\Delta\hat{\boldsymbol{\alpha}}_k^T\quad \Delta\hat{\mathbf{b}}_{g,k}^T\right]$$

$$\hat{\mathbf{q}}_k = \hat{\mathbf{q}}_{k|k-1} + 0.5\,\Xi(\hat{\mathbf{q}}_{k|k-1})\Delta\hat{\boldsymbol{\alpha}}_k$$

$$\hat{\mathbf{b}}_{g,k} = \hat{\mathbf{b}}_{g,k|k-1} + \Delta\hat{\mathbf{b}}_{g,k}$$

**Propagation**

$$\hat{\boldsymbol{\omega}}_{ib}^b(t) = \tilde{\boldsymbol{\omega}}_{ib}^b(t) - \hat{\mathbf{b}}_g(t)$$

$$\dot{\hat{\mathbf{q}}}_{b_0}^b(t) = \frac{1}{2}\Xi\left[\hat{\mathbf{q}}_{b_0}^b(t)\right]\hat{\boldsymbol{\omega}}_{ib}^b(t)$$

$$\dot{\mathbf{P}}(t) = \mathbf{F}(t)\mathbf{P}(t) + \mathbf{P}(t)\mathbf{F}^T(t) + \mathbf{G}(t)\mathbf{Q}(t)\mathbf{G}^T(t)$$

$$\mathbf{F}(t) = \begin{bmatrix} -[\hat{\boldsymbol{\omega}}_{ib}^b(t)\times] & -\mathbf{I}_{3\times 3} \\ \mathbf{0}_{3\times 3} & \mathbf{0}_{3\times 3} \end{bmatrix},\ \mathbf{G}(t) = \begin{bmatrix} -\mathbf{I}_{3\times 3} & \mathbf{0}_{3\times 3} \\ \mathbf{0}_{3\times 3} & \mathbf{I}_{3\times 3} \end{bmatrix}$$

---

In **Algorithm 1**, $\Xi[\mathbf{q}]$ is defined as

$$\Xi[\mathbf{q}] = \begin{bmatrix} \eta\mathbf{I}_{3\times 3} + (\boldsymbol{\rho}\times) \\ -\boldsymbol{\rho}^T \end{bmatrix} \tag{25}$$

**Algorithm 1** is a continues-discrete form of MEKF. The explicit discrete form of MEKF for on-board application is not presented here for brevity. The readers can refer to [12] for more information about this algorithm.

## IV. Conclusions

This note has traced the thread of optimization-based approach for SINS through exploring a new technique for joint estimation of attitude and inertial sensors drift biases. The joint estimation problem is posed as an interlaced procedure which contains dynamic attitude estimation and online optimization. The outputs of the two procedures are the necessary inputs for each other to construct the corresponding vector observations.

## V. My Comments

Initial alignment has been one of the difficult and attractive topics for SINS. To design practical initial alignment method, one should master SINS calculation, error analysis, state estimation and so on. If one has mastered initial alignment, he can also master inertial navigation technologies, so to speak. The initial alignment is the starting stage of SINS and studying of initial alignment can be viewed as the first step of studying inertial navigation technologies. This can be viewed as the reasons why so many papers related to initial alignment has been published. Unfortunately, most of these papers are only suitable for theoretical investigation and there is still a long way from becoming practical, however. Taking the nonlinear filtering based alignment for example, it is expected to unify the coarse and fine alignment stages through modeling nonlinear errors equations for SINS. However, the initial state covariance is a crucial factor for the attitude alignment performance. If the covariance is too small, the filtering may be diverging while if it is too big, the filtering steady state may be too large, which can not research the same accuracy level as the traditional two-stage method. The appropriate covariance setting is corresponding to the initial attitude misalignment which unfortunately, is unknown for initial alignment. In this respect, I don't prefer to make use of nonlinear filtering based alignment method in practice.

The OBA method is an attractive method for SINS attitude alignment and it has been applied in practice to the best of my knowledge. The aforementioned applications actually refer in particular to the original form of OBA in [5, 8]. The subsequent "modification" or "extension" are also limited to theoretical investigation. For the presented method in this note, I have not carried out the numerical and experimental tests. This is because that I know that the corresponding results will be not so satisfactory as expected, which is due to the coupling of errors sources. Actually, the method in [4] is also capable of estimating all the involved parameters theoretically. The unsatisfactory gyroscopes drift biases estimate is also due to the coupling of errors sources. The superiority of attitude alignment in [4] compared with traditional EKF based method is just derived through the optimization procedure which contains a special vector observations construction procedure. This superiority has also been demonstrated in our attitude estimation based alignment method [10]. Based on the aforementioned discussion, my choice of initial alignment in practice is making use of OBA method in [5, 8] to perform coarse alignment and making use of Kalman filtering to perform fine alignment.

## Acknowledgment

The author would like to thank Prof. Yuanxin Wu and Dr. Yulong Huang for the discussions on the SINS initial alignment problem.